# Concept Representation Learning with Contrastive Self-Supervised Learning

Daniel T. Chang (张遵)

*IBM (Retired)* dtchang43@gmail.com

**Abstract:** Concept-oriented deep learning (CODL) is a general approach to meet the future challenges for deep learning: (1) learning with little or no external supervision, (2) coping with test examples that come from a different distribution than the training examples, and (3) integrating deep learning with symbolic AI. In CODL, as in human learning, concept representations are learned based on concept exemplars. Contrastive self-supervised learning (CSSL) provides a promising approach to do so, since it: (1) uses data-driven associations, to get away from semantic labels, (2) supports incremental and continual learning, to get away from (large) fixed datasets, and (3) accommodates emergent objectives, to get away from fixed objectives (tasks). We discuss major aspects of concept representation learning using CSSL. These include dual-level concept representations, CSSL for feature representations, exemplar similarity measures and self-supervised relational reasoning, incremental and continual CSSL, and contrastive self-supervised concept (class) incremental learning. The discussion leverages recent findings from cognitive neural science and CSSL.

## 1 Introduction

The origins, recent advances and future challenges of *deep learning* are discussed in [1]. It points out that there are fundamental deficiencies of current deep learning that cannot be overcome by scaling alone. It suggests several directions for improvement:

1. Supervised learning requires too much labeled data. Humans are able to generalize well with far less experience.
2. Current deep learning systems are not robust to changes in distribution. Humans can quickly adapt to such changes with very few examples.
3. Current deep learning is most successful at perception tasks. Humans can perform higher-level cognition.

Specifically, the *future challenges for deep learning* involve (1) learning with little or no external supervision, (2) coping with test examples that come from a different distribution than the training examples, and (3) integrating deep learning with symbolic AI.

*Concepts* are the foundation of human learning, understanding, and knowledge integration and transfer [2]. Thus, to meet the future challenges for deep learning, it is crucial to explicitly incorporate concepts as fundamental tenets in deep learning.

We previously proposed *concept-oriented deep learning (CODL)* [2] as a general approach to do so. CODL extends current deep learning (rote feature representation learning) with *concept representation learning* and *conceptual understanding capability*. CODL addresses some of the major limitations of current deep learning: interpretability, transferability, contextual adaptation, and requirement for lots of labeled training data. The major aspects of CODL include concept graph, concept representations, concept exemplars, and concept representation learning systems supporting incremental and continual learning.

In CODL, as in human learning, *concept representations* are learned based on *concept exemplars* [2] which are *few typical instances of a concept*. This is feasible due to the high-level, semantically-segmented nature of concepts, in contrast to the low-level, semantically-nonrepresentative nature of features. Significantly, it avoids the need to gather and create large datasets, which can be difficult or costly, to use for training. *Concept representation learning* can be done using *supervised learning* or *unsupervised learning*. In the former case, the concept exemplars are labeled with the concepts (and possibly their instances and attributes). In the latter case, the concept exemplars are unlabeled. In both cases, *concept-invariant transformations* can be applied to concept exemplars to improve training.

Unsupervised representation learning is nowadays called *self-supervised learning*, which highlights the fact that the learning leverages the (unlabeled) data's inherent co-occurrence relationships as the self-supervision. It uses *pretext tasks* to learn a *general and transferrable feature representation*. There are two main approaches to self-supervised learning [4]:

- *Generative*: train an encoder to encode input x into a latent vector z and a decoder to *reconstruct x from z*.
- *Contrastive*: train an encoder to encode input x into a latent vector z to *measure similarity*.

The most popular generative model is the *autoencoder model* with many variants. The *Tiered Graph Autoencoder* model [3], which we previously proposed, is a variant of the autoencoder model.

*Contrastive self-supervised learning (CSSL)* [5] aims at *embedding* augmented versions of the same sample (i.e., *positive examples*) close to each other while pushing away embeddings from different samples (i.e., *negative examples*). The three most common components of CSSL are *encoders*, *pretext tasks* and *contrastive loss functions*. Encoders are responsible for mapping the input samples to a latent space of *embeddings*. To train an encoder, a pretext task is used that utilizes contrastive loss for backpropagation. For images, two of the most common pretext tasks are color transformation and geometric transformation. Contrastive loss functions are generally based on *similarity metrics* which measure the closeness between the embeddings of two samples. The trained *neural network backbone* can be used in downstream tasks such as classification and sample retrieval.

The promises and objectives of CSSL are [6]:

- *Data-driven associations*: to get away from semantic categories (labels)
- *Incremental and continual learning*: to get away from (large) fixed datasets
- *Emergent objectives*: to get away from fixed objectives (tasks)

These match very well with those of CODL. In CODL, concept representations can be learned using unsupervised learning; CODL supports incremental and continual learning; and concept representations are generic and can be used for various tasks.

In this paper, we discuss major aspects of *concept representation learning* using *CSSL*. The discussion leverages recent findings from *cognitive neural science* and *CSSL*.

## 2 Dual-Level Concept Representations

*Human brain* has traditionally been assumed to represent knowledge through the *embodiment of sensory experiences*. Recent behavioral and neuroimaging studies of visual knowledge with and without sensory experience [7], however, provide empirical evidence for the *neural coding of non-sensory, language-derived knowledge*, along with *sensory-derived representation*, in different brain systems. In particular, studies of neural activity patterns in congenitally blind and sighted groups have provided positive evidence for neural coding of non-sensory knowledge representation (for at least visual concepts) in dorsal ATL in the human brain, while also confirming the sensory-derived representation in the visual cortex in the sighted. Importantly, such coding of visual knowledge in dorsal ATL occurs not only in the blind but also the sighted brain, indicating that *even when sensory-derived properties are available there are both sensory-derived and non-sensory representations*. Other studies have found that *abstract concepts* (i.e., those without specific sensory referents, such as 'justice'), compared with *concrete concepts* (e.g., 'dog'), elicit stronger neural activation in the left ATL along with other regions encompassing the *language network*.

The *dual-coding neural knowledge representation framework* [7] is thus proposed as nature's solution to the challenges of knowledge representation. It consists of *embodied, sensory-derived knowledge representations*, from sensory experience, and *symbolic*, *language-derived knowledge representations*, from language experience. Sensor-derived representations are *modality specific* (visual, auditory, tactile, gustatory, olfactory, etc.); language-derived representations, on the other hand, are *amodal*. For concrete concepts, the two types of representations are *associated*.

The influential *hub-and-spoke model* [8] is similar to the dual-coding framework. The model consists of several sets of *spoke* units representing sensory and verbal elements of experience. There are also a set of hidden units (the *hub*) which do not receive external inputs but instead mediate between the various spokes. The spokes are *modality specific*, encode different information sources, and are integrated in the ventral-lateral ATL hub. The hub is the home of *amodal, symbolic representations*. The semantic hub and its modality-specific semantic white matter connections were recently *identified* [9] based on evidence from semantic dementia. The results show that the *hub region* works in concert with *nine other regions* in the semantic neural network for general semantic processing.

For *concept representations*, therefore, we use a *dual-level model*: the *embodied level* consists of *concept-oriented feature representations* [2], and the *symbolic level* consists of *concepts* in the form of a *concept graph* [2]. The symbolic level may be linked to Microsoft Concept Graph, or something comparable, which serves as the common / background conceptual knowledge base and the framework for conceptual understanding. For *concrete concepts*, the two levels are *associated / connected*. The embodied level corresponds to sensory-derived knowledge representations of the dual-coding framework; the symbolic level corresponds to language-derived knowledge representations of that framework.

In this paper, we focus our discussion on learning the embodied level of concept representations (i.e., *concept-oriented feature representations*) using *CSSL*.

## 3 CSSL for Feature Representations

*Visual encoding models* [10] are computational models for understanding how information is processed along the visual stream. In *humans*, visual information is processed by a cascade of neural computations and the mapping from the input stimulus space to the brain activity space is nonlinear. A *feature space* is usually introduced to assist model building, which assumes that the nonlinear mapping from the input space to the activity space is entirely contained by the *nonlinear mapping from the input space to the feature space*, such that only *linear mapping* is required *from the feature space to the activity space*. The hierarchical information processing mechanism of *deep neural networks* is highly similar to that of the visual cortex. Hence, visual encoding models based on *features extracted* using deep neural networks have been extensively studied. In particular, *CSSL* proves to be an effective method to extract brain-like *feature representations*.

The CSSL model [10] includes four modules: (1) *data augmentation* module, (2) *encoder* module, (3) *feature representation extraction* module, and (4) *contrastive loss function* module. It is based on SimCLRv2, a simple framework for contrastive learning of visual representations (see SimCLR in 6.2 Contrastive Self-Supervised CIL (CSS-CIL)). SimCLRv2 learns visual feature representations by maximizing the consistency between different views of the same sample (i.e., *positive examples*) and the distance between *negative examples* and *anchor points*. The anchor points, positive examples, and negative examples are created through data augmentation (flip, rotation, color distortion, etc.). The encoder is based on the ResNet50 encoder network. The deep convolutional encoder network can extract low-level, intermediate-level, and high-level features. The NT-Xent loss function is used to calculate the contrastive loss.

To evaluate the CSSL model's ability of characterizing brain representations [10], a *representation dissimilarity matrix (RDM)* is used to describe the model or brain representations, which is calculated as the *correlation distance (1 – Pearson correlation coefficient r)* between all pairs of model or brain representations. *Kendall rank correlation coefficient* is used to evaluate the similarity between a model RDM and a brain RDM. The results suggest that the CSSL model is a strong contender for explaining ventral stream visual representations. They also prove that *CSSL is an effective learning method to extract useful information from samples*. In fact, the *contrast mechanism* is an important mechanism of human learning.

For learning *concept-oriented feature representations* using CSSL, *concept exemplars* are used as input. The exemplars may be *labeled* or *unlabeled*. In the former case, the exemplars are labeled with the concepts (and possibly their instances and attributes). It is important to note that, unlike supervised learning, the labels are not used as targets in training. They are used to identify the learned feature representations in the embodied level as well as to associate these with concepts in the symbolic level. If the exemplars are unlabeled, *pseudo labels* are generated for candidate concepts, same as in human learning.

## 4 Exemplar Similarity Measures and Self-Supervised Relational Reasoning

For learning *concept-oriented feature representations* using *concept exemplars* as input, the commonly-used contrastive loss functions [5] are feature-oriented and not suitable for contrasting concepts. We need *concept-oriented loss functions* based on, e.g., exemplar similarity measures and self-supervised relational reasoning, which are discussed below.

### 4.1 Exemplar Similarity Measures

In *human learning*, each interaction we have with a *concept* (e.g., "dog" or "justice") is unique. To form a high-level representation of a concept, we must draw on *similarities between (concept) exemplars* to form new or updated conceptual knowledge. Two *exemplar (pattern) similarity measures* — *pattern robustness* and *encoding-retrieval similarity* — are particularly important [11]. Pattern robustness indicates that a concept is robustly represented relative to other concepts; encoding-retrieval similarity positively predicts subsequent memory performance. Exemplar similarity is an important predictor of memory for novel concept-exemplar pairings even when the concept includes multiple exemplars. Importantly, established predictive relationships between exemplar similarity and subsequent memory do not require visually identical stimuli (i.e., are not simply due to low-level visual overlap between exemplars).

Any object or entity can be thought of as being in a set of hierarchically organized concepts, i.e., a *concept taxonomy* [2]. In a concept taxonomy, the concepts that are higher in the hierarchy (e.g., bird) are *superordinate* to the lower-level concepts; the lower-level concepts (e.g., sparrow) are *subordinate* to the higher-level ones. A person's ability to access a learned concept through these levels is important for memory, with some levels being remembered (e.g., basic levels [2]), while others are not. *Pattern robustness* [11] is defined as the difference between *within-concept* and *between-concept* similarities. For *superordinates* (e.g., bird), the within-concept similarity is calculated for each superordinate level as the *Pearson correlation coefficients* between matching superordinate level exemplars. The values are Fisher-z corrected and averaged across all superordinate levels. The between-concept similarity is calculated in the same manner but across taxonomical groups (e.g., bird with non-bird groups). For *subordinates* (e.g., dove, sparrow), the calculation is done similarly. However, the between-concept similarity is always calculated between subordinates within the same superordinate level (e.g., bird).

### 4.2 Self-Supervised Relational Reasoning

An important factor in *human learning* is the acquisition of new knowledge by *relating concepts*. Relationships among concepts allow a person to ignore irrelevant perceptual features and focus on *relevant non-obvious properties*. *Self-supervised relational reasoning* [12] is proposed to exploit a similar mechanism in *CSSL*. It can be used as a *pretext task* to build feature representations in a *neural network backbone*, by training a *relation head* on unlabeled data to discriminate how concepts relate to themselves (*intra-reasoning*) and other concepts (*inter-reasoning*). Once the system (backbone + relation head) has been trained, the relation head is discarded, and the backbone is used in downstream tasks (e.g. classification, sample retrieval).

*Intra-reasoning* consists of coupling two random augmentations of the same exemplar $\{\mathbb{A}(\epsilon_i), \mathbb{A}(\epsilon_i)\} \rightarrow$ "same" (*positive pair*), whereas *inter-reasoning* consists of coupling two random exemplars $\{\mathbb{A}(\epsilon_i), \mathbb{A}(\epsilon_j); i \neq j\} \rightarrow$ "different" (*negative pair*). An example is coupling two different views of the same apple to build the positive pair, and coupling an apple with a different fruit to build the negative pair. Intra-reasoning is thus *within-concept*, and inter-reasoning is *between-concept*.

The *learning objective* [12] consists of an intra-reasoning loss term and an inter-reasoning loss term. The *intra-reasoning loss term* is the sum of the differences between the *relation score* of the encodings of each *positive pair* and a target value of 1 (perfect correlation); the *inter-reasoning loss term* is the sum of the differences between the relation score of the encodings of each *negative pair* and a target value of 0 (perfect independence). The relation score is a learnable, probabilistic similarity measure (e.g. exemplar similarity measures). By *minimizing* the learning objective, each feature representation (encoding) is pushed towards a positive neighborhood (intra-reasoning) and repelled from a complementary set of negative neighborhoods (inter-reasoning).

### 4.3 CSSL Using Exemplars as Input

In *human learning*, learning a concept involves few exemplars and, further, the exemplars only need sufficient details to be distinguishable from exemplars of different concepts. This is in great contrast to current deep learning which requires large amount of data with great details. *CSSL using exemplars as input*, therefore, involves *small datasets* and requires only *data with sufficient details* that are suitable for use in exemplar similarity measures and self-supervised relational reasoning.

*S3L* [13] proposes a learning paradigm for CSSL that includes three parts: *small dataset*, *small resolution*, and *small model*. Various experiments show that the paradigm achieves significant benefits with small training cost, especially on small datasets. CSSL using exemplars as input matches this learning paradigm.

## 5 Incremental and Continual CSSL

In *human learning*, we learn concepts incrementally, with or without having existing concepts for comparison. And we continue to learn new concepts and new aspects of existing concepts. When presented with new concepts to learn, we leverage knowledge from previous concepts and integrate newly learned knowledge into previous concepts. Therefore, as discussed in [2], *incremental and continual learning* is a critical aspect of CODL. The main challenge for *CSSL* w.r.t. incremental and continual learning is the same as that for incremental and continual learning models in general. They suffer from *catastrophic forgetting*, i.e., training a model with data for new concepts and updating its parameters interferes with previously learned concepts. This leads to a drastic drop in performance on previously learned concepts [14]. As the number of incremental steps goes up, the performance degradation increases.

*Continual Contrastive Self-supervised Learning (CCSL)* [14] is designed to alleviate the catastrophic forgetting problem in CSSL. First, it uses a *rehearsal method* which keeps a few *exemplars* from the previous data. Previous data are projected into feature embedding space and K-Mean algorithm is used to group the feature vectors by similarity. For each group, it stores the most consistent samples (i.e., exemplars) by measuring their feature vector variance. Second, instead of directly combining saved exemplars with the current dataset for training, it leverages *self-supervised knowledge distillation* to transfer contrastive information among previous data to the current network by mimicking *similarity score distribution* inferred by the old network over a set of saved exemplars. This strengthens the *intra-contrast* among previous data. Finally, note that when the network learns feature presentations of current data, the region of feature vectors of current data in embedding space may mix up with the region of feature vectors of previous data due to the lack of contrast between previous data and current data (called *inter-contrast*). Therefore, it builds an *extra sample queue* to assist the network to distinguish between previous and current data and prevent mutual interference while learning their own feature representations.

CCSL adapts *MoCo (Momentum Contrast)* [14-15] as the base model. MoCo treats contrastive learning as *dictionary look-up* and builds a dynamic dictionary with a queue and a moving-average encoder. This enables building a large and consistent dictionary on-the-fly that facilitates CSSL. MoCo contains a *query encoder $f_q(.)$*, a *key encoder $f_k(.)$*, and a *memory bank*. An *input x* is first transformed to *two views $x_q$ and $x_k$*. Then, $x_q$ and $x_k$ are fed into the encoders, $f_q$ and $f_k$, respectively to get the *query q* and the *positive key sample $k_+$*. The memory bank stores *negative key samples $\{k_1, k_2, …, k_n\}$* of q. MoCo adopts a *contrastive loss* whose value is low when q is similar to the positive key $k_+$ and dissimilar to all negative keys $k_i$. Furthermore, the *query encoder parameters $\theta_q$* are updated by back-propagation and the *key encoder parameters $\theta_k$* are updated by momentum strategy, respectively.

# 6 Contrastive Self-Supervised Concept (Class) Incremental Learning

For concrete things (*objects*), 'concept' is usually referred to as '*class*'; for abstract things (*entities*), 'concept' is commonly referred to as '*type*' [2]. Our discussion in the following focuses on concrete concepts. Therefore, we use the terms "concept" and "class" interchangeably.

*Class incremental learning (CIL)* [16-17] is a special case of incremental (and continual) learning, where each incremental data consists of new class samples. *Contrastive self-supervised CIL (CSS-CIL)* provides an alternative approach to CCSL in alleviating the catastrophic forgetting problem in CSSL. CCSL add incremental and continual learning support to base CSSL models (e.g., MoCo); CSS-CIL, on the other hand, adds CSSL support to base CIL models (e.g., exemplar-based methods, as discussed below).

## 6.1 Class Incremental Learning (CIL)

A *CIL problem T* consists of a sequence of *N tasks* [16]. A task is a set of *classes* disjoint from classes in other (previous or future) tasks. Formally, T can be defined as:

$$T = [(C^1, D^1), (C^2, D^2), \ldots, (C^N, D^N)],$$

where each *task t* is represented by a set of classes $C^t = \{c^t_1, c^t_2, \ldots, c^t_{N^{\wedge}t}\}$ and training data $D^t$. During training for task t, the *incremental learner* only has access to $D^t$, and the tasks do not overlap in classes (i.e. $C^i \cap C^j = 0$ if $i \neq j$).

Traditionally, the objective of CIL is to learn a *unified classifier* from a sequence of data from different classes, via *supervised learning*, in which case [16]:

$$D^t = \{(\mathbf{x}_1, y_1), (\mathbf{x}_2, y_2), \ldots, (\mathbf{x}_{m^{\wedge}t}, y_{m^{\wedge}t})\},$$

where **x** are input features for a training sample, and y is the label corresponding to **x**. The incremental learner is a deep neural network parameterized by weights $\theta$:

$$\mathbf{o}(\mathbf{x}) = h(\mathbf{x}; \theta).$$

It is common to split the deep neural network into a *feature extractor f(.)* with weights $\varphi$ and a *linear classifier g(.)* with weights $\nu$:

$$\mathbf{o}(\mathbf{x}) = g(f(\mathbf{x}; \varphi); \nu).$$

This assumes all the nonlinearity of the deep neural network resides in the feature extractor. We note that the same is usually assumed in visual encoding models (see 3 CSSL for Feature Representations).

There are two main categories of CIL methods [16]: *exemplar-based (rehearsal) methods* that store a limited set of exemplars to prevent forgetting of previous classes; *regularization-based methods* that aim to minimize the impact of learning new classes on the weights that are important for previous classes. We focus on exemplar-based methods. The three crucial components of an exemplar-based method [17] include a *memory buffer* to store few exemplars from old classes (e.g., using *herding heuristics*), a *forgetting constraint* (e.g., *knowledge distillation*) to keep previous knowledge while learning new classes, and a *learning system* that balances old and new classes. iCaRL [18] is the most typical exemplar-based method and employs knowledge distillation as forgetting constraint. We note that CCSL adopts the exemplar-based method (see 5 Incremental and Continual CSSL).

## 6.2 Contrastive Self-Supervised CIL (CSS-CIL)

*Self-Supervised Class Incremental Learning (SSCIL)* [19] is a framework for CSS-CIL. It discards the classifier and data labels, as these are used in supervised learning, of a *CIL model* (e.g., iCaRL), and provides three different *class incremental schemes* to *simulate CIL*: random class scheme, semantic class scheme, and clustering scheme. SSCIL adopts *SimCLR* for CSSL support, and uses *Linear Evaluation Protocol (LEP)* and *Generalization Evaluation Protocol (GEP)* to evaluate the model's classification ability and robustness.

The *random class scheme* splits a dataset D into N class-sets $\{D^1, D^2, …, D^N\}$ based on random classes. The number of classes in each class-set is equal and the class-sets do not intersect. Nevertheless, the semantic space of each class-set is likely to have a large amount of overlap. For example, although 'cobra' and 'rattlesnake' are different classes, they all belong to the 'snake' family. As a result, the catastrophic forgetting of classes is concealed because of the similarity of semantic information between class-sets. In order to avoid the overlap of semantics between classes in each training phase, the *semantic class scheme* defines each class according to the WordNet and split the dataset D into N class-sets where the samples in the same class-set $D^i$ have similar semantic information while the samples in different class-sets $\{D^i, D^j; i \neq j\}$ have few overlapping in semantic space. Finally, the *clustering scheme* uses K-means clustering to divide the dataset D into N class-sets. The number of samples in each class-set is equal, but there are many intersections between the classes of each class-set.

*SimCLR* [20] is a simple framework for contrastive learning of visual representations. It learns feature representations by maximizing agreement between differently *augmented views* of the same sample via a *contrastive loss in the latent space*. SimCLR consists of four major components: (1) a *data augmentation* $\mathcal{A}$ module that transforms any given sample randomly resulting in two correlated views (i.e., a *positive pair*) of the same sample, (2) a neural network *feature extractor f(.)* that extracts *feature representations* from augmented samples, (3) a neural network *projection head g(.)* that maps feature representations to the *latent space* where contrastive loss is applied, and (4) a *contrastive loss function* $\mathcal{L}$ defined for a contrastive prediction task that identifies *positive pairs*.

To use the evaluation protocols, the feature extractor $f^t(.)$ is frozen after the training in phase t has completed. The *Linear Evaluation Protocol (LEP)* uses all of the class-sets $\{D^1, D^2, …, D^t\}$ trained thus far to fit a linear classifier $M^t(.)$ on top of the $f^t(.)$. The test accuracy

$$LEP^t = M^t(f^t(D^t_{test}))$$

indicates the model's ability to classify all known classes at the phase t, where $D^t_{test}$ is the test set of the class-sets $\{D^1, D^2, …, D^t\}$. The *Generalization Evaluation Protocol (GEP)*, on the other hand, uses the full dataset D to fine-tune a linear classifier $M^t(.)$ on top of the $f^t(.)$. The test accuracy

$$GEP^t = M^t(f^t(D_{test}))$$

indicates how much knowledge the model has at the phase t, where the $D_{test}$ is the test set of the full-dataset D.

## 7 Conclusion

Based on recent findings from cognitive neural science and contrastive self-supervised learning (CSSL), we find: (1) learning concept representations, which are general and transferable, is crucial to meet the future challenges for deep learning, and (2) CSSL provides a promising approach to learn concept representations at the embodied level (i.e., concept-oriented feature representations) based on concept exemplars. We discuss major aspects of concept representation learning using CSSL, particularly incremental and continual learning since it is infeasible and unrealistic to learn all concepts and their key attributes at once. For future work, we plan to explore exemplar-based concept (class) incremental learning with contrastive self-supervised relational reasoning and exemplar similarity measures.

**Acknowledgement:** Thanks to my wife Hedy (郑期芳) for her support.